\pdfoutput=1
\documentclass{article}
\usepackage{spconf,amsmath,graphicx}

\title{Interpretable Multimodal MRI--Clinical Fusion for Functional Outcome Prediction After Acute Ischemic Stroke}

\name{Sina Raeisadigh$^{1,4}$, Henning Müller$^{2,5}$, Abderrahmane Hedjoudje$^{3}$}

\address{$^{1}$ Department of Computer Science, University of Geneva, Switzerland \\
         $^{2}$ Service of Medical Informatics, University Hospital of Geneva, Switzerland \\
         $^{3}$ Department of Imaging and Medical Informatics, University of Geneva, Switzerland \\
         $^{4}$ Informatics Institute, University of Applied Sciences Western Switzerland (HES-SO), Sierre, Switzerland \\
         $^{5}$ The Sense, Lausanne and Sion, Switzerland}
\begin{document}

\maketitle
\begin{abstract}
This study evaluates the added prognostic value of day-1 multimodal MRI for predicting three-month functional outcome after acute ischemic stroke (AIS). Sixty-six AIS patients with ADC, DWI, FLAIR, and clinical data were analyzed. Three-dimensional pretrained ResNet embeddings were extracted from each MRI modality, reduced using fold-isolated principal component analysis, and combined with structured clinical variables and ADC-derived lesion biomarkers. Linear support vector machines were evaluated using eight-fold stratified group cross-validation across MRI modality combinations. Clinical-only modeling achieved strong performance (AUC = $0.899 \pm 0.151$), while MRI-only models showed limited discrimination (AUC $\approx$ $0.67$--$0.73$). Among multimodal configurations, FLAIR-clinical fusion achieved the highest mean cross-validated performance (AUC = $0.948 \pm 0.061$), followed by ADC-clinical fusion (AUC = $0.941 \pm 0.094$). Full ADC-DWI-FLAIR fusion did not consistently outperform single-modality clinical fusion. Feature-importance analysis identified age, NIHSS subscores, ADC lesion burden, and learned FLAIR representations as key contributors, while qualitative FLAIR activation maps showed responses over lesion-like regions. These findings suggest that compact day-1 FLAIR and ADC representations provide complementary prognostic information beyond clinical variables and support interpretable, data-efficient outcome prediction in AIS.
\end{abstract}

\begin{keywords}
acute ischemic stroke, multimodal MRI, clinical-imaging fusion, functional outcome prediction, modified Rankin Scale, explainable AI
\end{keywords}
\section{Introduction}

Acute ischemic stroke (AIS) remains a major cause of adult disability and long-term functional dependence worldwide~\cite{feigin2021global}. Early prediction of three-month functional outcome is clinically important because it can support treatment planning, rehabilitation allocation, patient counseling, and trial stratification. The modified Rankin Scale (mRS) is widely used as the standard functional outcome endpoint after stroke, with poor outcome commonly defined as mRS $>$ 1 or mRS $>$ 2 depending on the clinical setting~\cite{banks2007stroke}. However, accurate individualized prognosis remains challenging because recovery depends on complex interactions between baseline neurological severity, age, vascular risk factors, lesion burden, treatment response, and early tissue evolution.

Clinical variables such as age, pre-stroke disability, and the National Institutes of Health Stroke Scale (NIHSS) are among the strongest predictors of post-stroke outcome~\cite{wouters2018prediction, heiss2017neuroimaging}. Nevertheless, clinical scores alone do not directly capture lesion location, infarct extent, or tissue-level changes after reperfusion. Magnetic resonance imaging (MRI), especially diffusion-weighted imaging (DWI) and apparent diffusion coefficient (ADC) maps, provides sensitive markers of acute ischemic injury~\cite{warach1992fast}. Fluid-attenuated inversion recovery (FLAIR) imaging can provide complementary information related to lesion visibility, edema, and subacute tissue changes. Together, these modalities may improve outcome prediction by representing different aspects of ischemic tissue damage.

Recent machine learning and deep learning methods have shown promise for combining imaging and clinical data in stroke prognosis~\cite{nielsen2018prediction, kim2023deep, ma2024transformer}. Three-dimensional convolutional neural networks can extract volumetric representations from MRI without requiring handcrafted radiomic features, while linear classifiers can provide more transparent decision boundaries in small clinical cohorts~\cite{lecun2015deep, chen2019med3d, cardoso2022monai}. However, many studies either focus on a single imaging modality, use imaging as a black-box input, or do not explicitly separate the contribution of clinical variables, MRI modality choice, and lesion-derived biomarkers. This makes it difficult to determine whether performance gains come from deep image representations, structured clinical predictors, or explicit lesion burden measures.

A further challenge is data efficiency. Stroke imaging datasets with complete longitudinal clinical outcomes are often small, making end-to-end deep learning prone to overfitting. Hybrid approaches that use pretrained 3D networks as feature extractors, followed by fold-isolated dimensionality reduction and linear classification, may offer a practical compromise between representation learning and interpretability~\cite{wold1987principal, cortes1995support, pedregosa2011scikit_learn}. In this setting, systematic evaluation of MRI modality combinations is important: adding more modalities does not necessarily improve generalization, especially when sample size is limited and modality-specific features are redundant or noisy. Our previous work showed that day-1 ADC-based multimodal modeling improved three-month mRS prediction compared with baseline ADC imaging, highlighting the prognostic value of early post-treatment diffusion changes~\cite{raeisadigh2026comparing}. However, that study focused primarily on ADC and did not systematically evaluate the relative contribution of ADC, DWI, and FLAIR modality combinations or explicitly separate the effect of ADC-derived lesion biomarkers from MRI image inputs.

In this work, we evaluate day-1 multimodal MRI fusion for predicting three-month functional outcome after AIS in a quality-controlled cohort of 66 patients. We compare ADC, DWI, and FLAIR representations individually and in combination, with and without structured clinical variables and ADC-derived lesion biomarkers. We use pretrained 3D ResNet feature extraction, fold-isolated principal component analysis, and class-balanced linear support vector machines under eight-fold stratified group cross-validation. In addition to predictive performance, we provide interpretability through filtered linear coefficient analysis and qualitative FLAIR CNN activation maps. Our goal is to identify which imaging modalities and lesion-derived features add prognostic value beyond clinical variables, while maintaining a data-efficient and interpretable modeling framework.
\section{Methods}

\subsection{Cohort and Outcome}
This retrospective study included 66 patients with acute ischemic stroke (AIS) who had day-1 ADC, DWI, and FLAIR MRI available after quality control, together with structured clinical variables and three-month modified Rankin Scale (mRS) outcome. Clinical variables included demographics, vascular risk factors, pre-stroke disability, and NIHSS subscores at baseline and day 1. The binary endpoint was favorable outcome (mRS $\leq 1$) versus unfavorable outcome (mRS $>1$). Patient identifiers were used as grouping variables to prevent leakage across cross-validation folds.

\subsection{MRI Preprocessing}
MRI data were loaded from clinical DICOM series and processed in physical space. Candidate image series were identified from DICOM metadata, grouped by SeriesInstanceUID, and the largest valid image series was selected for each patient and modality. Slice ordering was checked using spatial DICOM information when available, and duplicate slice positions were removed. ADC, DWI, and FLAIR volumes were resampled with linear interpolation to a common physical spacing of $1.02 \times 1.02 \times 5.4$ mm and converted to depth-height-width order. Volumes were center-padded to $32 \times 256 \times 256$ without cropping anatomy. Patients were excluded if any requested modality failed prespecified physical coverage requirements. Per-volume robust intensity normalization was applied using finite, nonzero valid voxels while excluding padded background.

\subsection{Lesion Biomarkers and MRI Features}
Three ADC-derived lesion-burden features were used: lesions defined at ADC thresholds of $480 \times 10^{-6} \, \text{mm}^2/\text{s}$ and $620 \times 10^{-6} \, \text{mm}^2/\text{s}$, and a combined weighted lesion measure incorporating both thresholds. Small isolated connected components were removed before feature calculation. MRI representations were extracted from ADC, DWI, and FLAIR using pretrained three-dimensional ResNet branches implemented in MONAI. Modality embeddings were evaluated individually and in all possible combinations to assess the relative contribution of ADC, DWI, FLAIR, and multimodal MRI fusion.
\subsection{Clinical-Imaging Fusion}
For each configuration, selected MRI embeddings were concatenated with the corresponding clinical/tabular feature block. We evaluated MRI-only models, clinical-only models, and MRI-clinical fusion models. To isolate the contribution of ADC-derived lesion biomarkers, clinical and MRI-clinical configurations were evaluated with and without these lesion features. The clinical-only model without ADC lesion biomarkers was used as the reference baseline for reporting incremental performance.

\subsection{Classification and Evaluation}
Dimensionality reduction and feature scaling were performed within each training fold and then applied to the held-out validation fold to prevent information leakage. Classification used a class-balanced linear support vector machine. Models were evaluated using eight-fold stratified group cross-validation with patient-level grouping. Performance was reported as mean $\pm$ standard deviation across validation folds using AUC, accuracy, and F1-score. Models were ranked by mean cross-validated AUC. Network and classifier settings were fixed across modality comparisons,
and all learned preprocessing was fitted within each training fold. The primary
FLAIR-clinical model with ADC-derived lesion biomarkers was compared with
the clinical-only baseline using an exact two-sided paired Wilcoxon signed-rank
test on validation AUCs from the corresponding cross-validation folds.
\begin{table*}[t]
\centering
\caption{Cross-validation performance for selected configurations. Values are mean $\pm$ standard deviation across folds. Delta AUC is computed relative to the clinical-only model without ADC lesion biomarkers.}
\label{tab:main_results}
\scriptsize
\setlength{\tabcolsep}{11pt}
\renewcommand{\arraystretch}{1.12}
\begin{tabular}{lccccc}
\hline
\textbf{Configuration} &
\textbf{Train AUC} &
\textbf{Val AUC} &
\textbf{Val Acc} &
\textbf{Val F1} &
\textbf{$\Delta$ Val AUC} \\
\hline

FLAIR + clinical + ADC lesions
& $0.949 \pm 0.018$
& $\mathbf{0.948 \pm 0.061}$
& $0.802 \pm 0.093$
& $0.732 \pm 0.175$
& $\mathbf{+0.049}$ \\

ADC + clinical + ADC lesions
& $0.941 \pm 0.012$
& $0.941 \pm 0.094$
& $0.816 \pm 0.097$
& $0.776 \pm 0.086$
& $+0.042$ \\

ADC + FLAIR + clinical + ADC lesions
& $0.959 \pm 0.006$
& $0.932 \pm 0.070$
& $0.788 \pm 0.105$
& $0.747 \pm 0.161$
& $+0.033$ \\

ADC + DWI + FLAIR + clinical + ADC lesions
& $0.963 \pm 0.011$
& $0.926 \pm 0.107$
& $0.708 \pm 0.130$
& $0.636 \pm 0.198$
& $+0.027$ \\

ADC + DWI + clinical + ADC lesions
& $0.956 \pm 0.014$
& $0.922 \pm 0.118$
& $0.755 \pm 0.118$
& $0.674 \pm 0.184$
& $+0.023$ \\

Clinical only, without ADC lesions
& $0.913 \pm 0.017$
& $0.899 \pm 0.151$
& $\mathbf{0.845 \pm 0.161}$
& $\mathbf{0.820 \pm 0.145}$
& $0.000$ \\

Clinical only, with ADC lesions
& $0.923 \pm 0.017$
& $0.890 \pm 0.147$
& $0.767 \pm 0.157$
& $0.702 \pm 0.185$
& $-0.010$ \\

FLAIR only
& $0.680 \pm 0.040$
& $0.719 \pm 0.235$
& $0.609 \pm 0.185$
& $0.547 \pm 0.264$
& $-0.180$ \\

DWI only
& $0.711 \pm 0.021$
& $0.706 \pm 0.199$
& $0.620 \pm 0.140$
& $0.544 \pm 0.083$
& $-0.193$ \\

ADC only
& $0.652 \pm 0.029$
& $0.675 \pm 0.220$
& $0.641 \pm 0.196$
& $0.565 \pm 0.309$
& $-0.224$ \\
\hline
\end{tabular}
\end{table*}

\subsection{Explainability}
Global interpretability was assessed using a direct linear probe on FLAIR embeddings, clinical variables, and ADC lesion biomarkers. Rare binary variables were excluded from displayed feature-importance plots to reduce coefficient instability. Feature importance was summarized from average linear coefficients across folds. Local imaging interpretation used qualitative intermediate FLAIR CNN activation maps, shown alone and overlaid on raw FLAIR slices. These maps were used as qualitative activation visualizations, not as lesion segmentation or causal attribution.
\section{Results}
\subsection{Overall Model Performance}
Table~\ref{tab:main_results} summarizes the main cross-validation results. The clinical-only baseline without ADC-derived lesion biomarkers achieved strong performance (AUC $=0.899 \pm 0.151$, accuracy $=0.845 \pm 0.161$, F1 $=0.820 \pm 0.145$). MRI-only models showed substantially lower discrimination, with mean AUCs ranging from $0.675$ to $0.719$, indicating that image embeddings alone were insufficient for robust prediction in this cohort.

The best-performing configuration was FLAIR combined with clinical variables and ADC-derived lesion biomarkers, achieving AUC $=0.948 \pm 0.061$, accuracy $=0.802 \pm 0.093$, and F1 $=0.732 \pm 0.175$. This represented a numerical AUC improvement of $+0.049$ over the clinical-only reference. The paired fold-level AUC
difference was significant according to an exact two-sided Wilcoxon signed-rank
test ($p=0.0078$). ADC-clinical fusion with ADC lesion biomarkers also performed strongly (AUC $=0.941 \pm 0.094$), followed by ADC-FLAIR clinical fusion with lesion biomarkers (AUC $=0.932 \pm 0.070$). In contrast, combining all three MRI modalities did not provide the highest performance, suggesting that full modality fusion may introduce redundant or noisy information in a small-sample setting.

\subsection{Contribution of MRI Modalities and ADC Lesion Biomarkers}
Among single-modality clinical-imaging fusion models, FLAIR and ADC provided the strongest complementary information. FLAIR-clinical fusion with ADC lesion biomarkers achieved the highest AUC, while ADC-clinical fusion achieved the highest F1 among the multimodal models. DWI-clinical fusion was weaker (AUC $=0.893 \pm 0.100$ with ADC lesion biomarkers), and DWI-FLAIR fusion did not improve performance over the clinical baseline.

ADC-derived lesion biomarkers improved several MRI-clinical configurations, particularly FLAIR-clinical fusion, which increased from AUC $=0.889 \pm 0.115$ without lesion biomarkers to $0.948 \pm 0.061$ with lesion biomarkers. However, lesion biomarkers did not improve the clinical-only model, which decreased from AUC $=0.899 \pm 0.151$ without lesion biomarkers to $0.890 \pm 0.147$ with lesion biomarkers. This suggests that ADC-derived lesion burden was most useful when combined with learned MRI representations rather than as a standalone addition to clinical variables.

\subsection{Explainability}
\begin{figure}[t]
    \centering
    \includegraphics[width=0.47\textwidth]{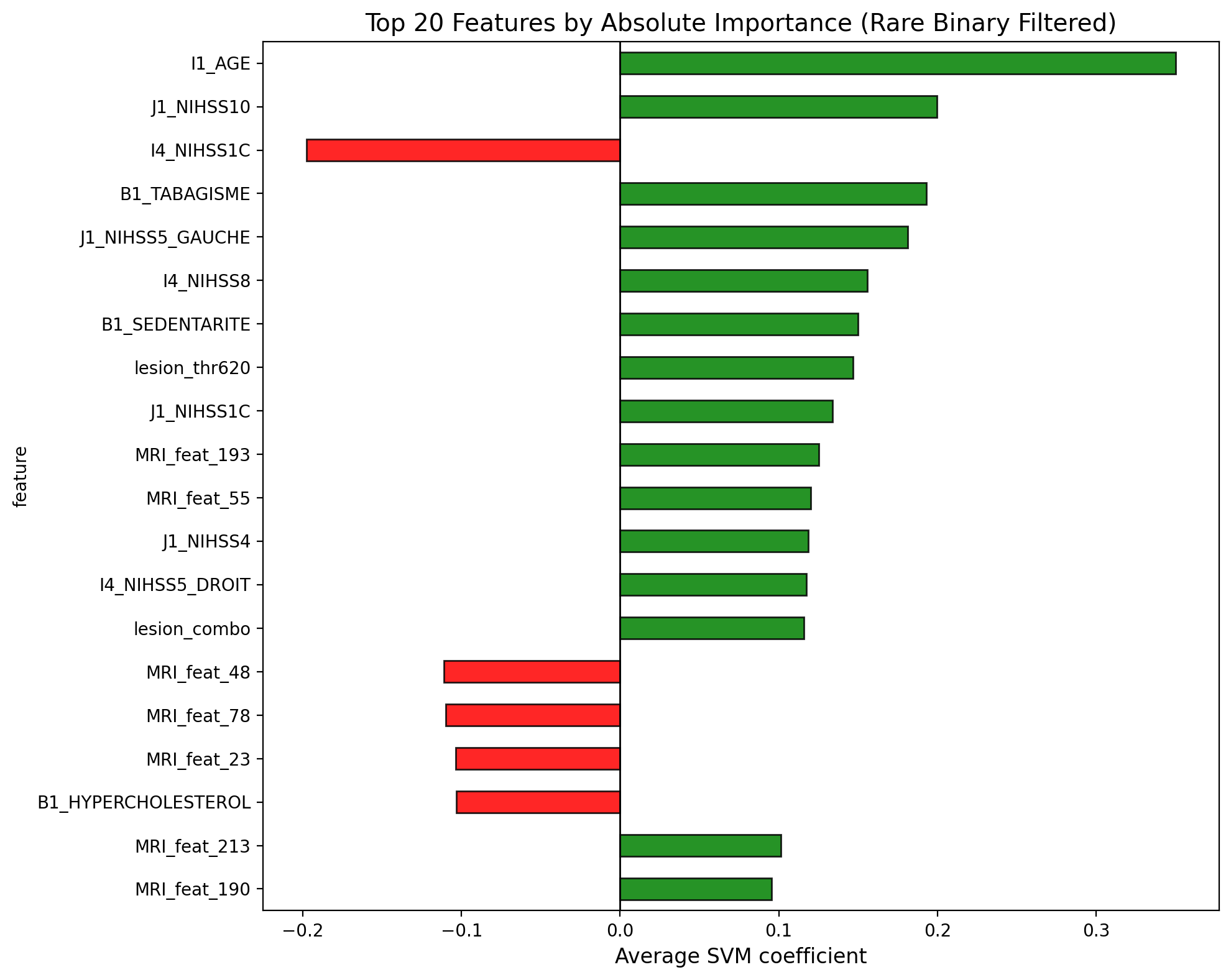}
    \caption{
    Filtered global feature importance from the direct linear probe. 
    Bars show average SVM coefficients across folds after excluding rare binary variables from visualization. 
    Positive coefficients indicate association with unfavorable outcome (mRS $>1$), while negative coefficients indicate association with favorable outcome. 
    Age, NIHSS-related variables, ADC lesion burden, and learned MRI features contributed to the prediction.
    }
    \label{fig:feature_importance}
\end{figure}

Filtered coefficient analysis identified age, NIHSS-related variables, ADC lesion burden, and learned MRI features among the most influential predictors (Fig.~\ref{fig:feature_importance}). Positive coefficients indicated association with unfavorable outcome, while negative coefficients indicated association with favorable outcome. After excluding rare binary variables from visualization, the most prominent contributors included age, day-1 NIHSS subscores, ADC lesion burden features, and several learned FLAIR-derived imaging features.

Qualitative activation maps showed spatial responses overlapping visible
FLAIR hyperintensities in the illustrated cases. The selected feature contributes to the FLAIR representation used by the final classifier, and its activation map indicates the image regions associated with that feature's response. However, these exploratory visualizations should not be interpreted as lesion segmentations or as evidence of a causal relationship between the highlighted regions and outcome.
\section{Discussion}

\begin{figure}[t]
    \centering
    \includegraphics[width=0.47\textwidth]{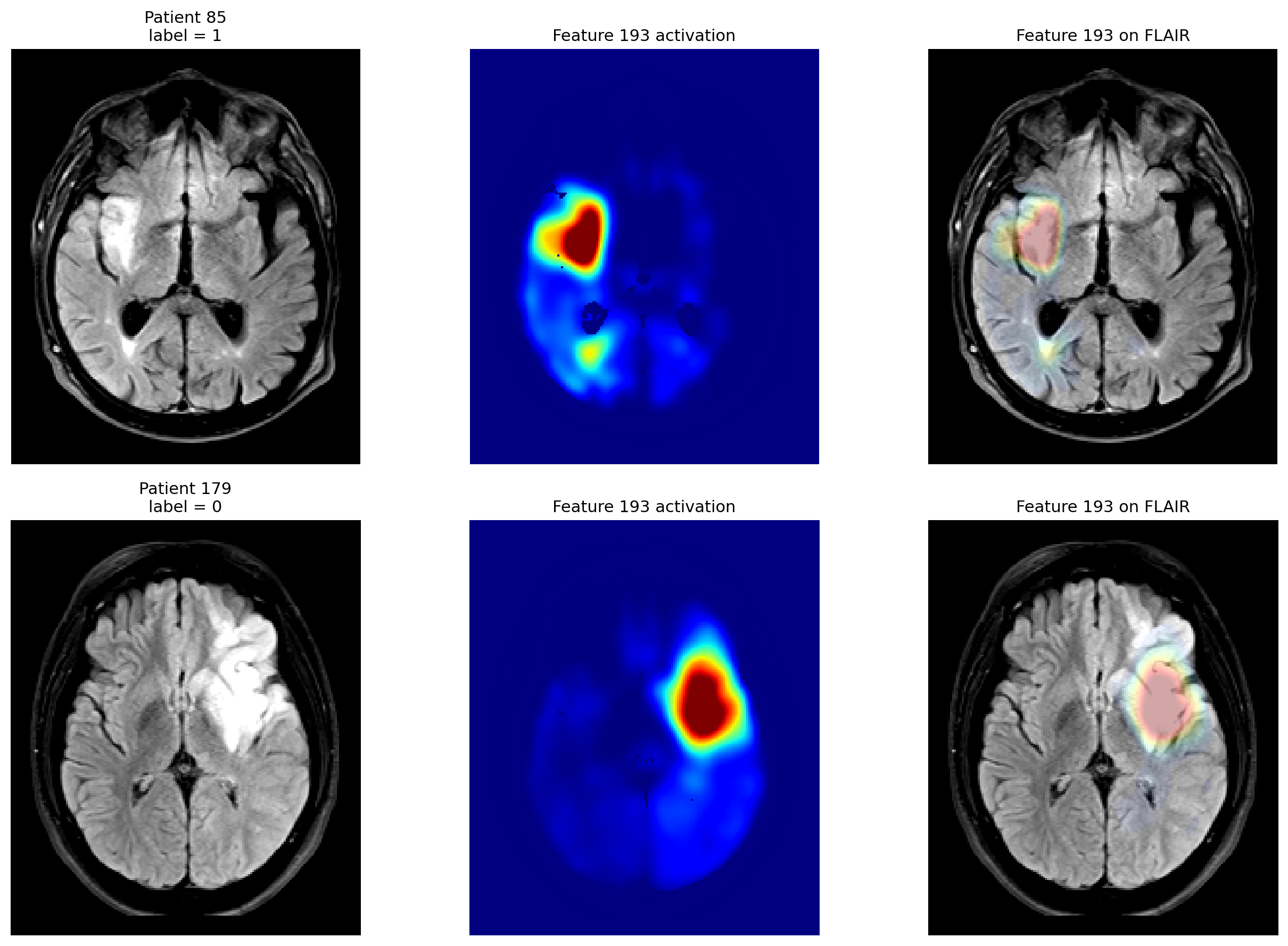}
    \caption{
    Qualitative FLAIR CNN activation visualization for representative patients. 
    Each row shows the raw FLAIR slice, the CNN activation map, and the activation overlaid on FLAIR. 
    The selected activation channel responds over lesion-like FLAIR hyperintense regions. 
    These maps are qualitative activation visualizations and should not be interpreted as lesion segmentation or causal attribution.
    }
    \label{fig:flair_activation}
\end{figure}
This study evaluated whether day-1 multimodal MRI adds prognostic information beyond structured clinical variables for predicting three-month functional outcome after AIS. FLAIR combined with clinical variables and ADC-derived lesion biomarkers achieved the highest cross-validated AUC, whereas the clinical-only model achieved the highest accuracy and F1-score. AUC was used as the primary ranking metric because it evaluates discrimination across decision thresholds, while accuracy and F1 depend on a single operating threshold. Thus, multimodal fusion improved overall patient ranking, although this advantage did not translate into better threshold-dependent classification.

The strong clinical-only performance is consistent with evidence that age, pre-stroke disability, and NIHSS are major predictors of post-stroke recovery~\cite{wouters2018prediction,heiss2017neuroimaging}. MRI-only models performed substantially worse, suggesting that image embeddings alone were insufficient in this small cohort. Nevertheless, FLAIR- and ADC-clinical fusion achieved higher AUCs when combined with clinical information and ADC lesion biomarkers, indicating that day-1 imaging provides complementary tissue-level information in the clinical context.

Full ADC-DWI-FLAIR fusion did not outperform the best single-modality clinical fusion model. This may reflect redundancy between modalities, limited sample size, and the introduction of noisy features, emphasizing the importance of controlled modality-combination experiments.

Explainability analysis identified age, NIHSS-related variables, ADC lesion burden, and learned MRI features among the most influential predictors. The selected FLAIR channel responded spatially to visible hyperintense abnormalities in the illustrated cases. These visualizations remain exploratory and should not be interpreted as lesion segmentation or causal attribution.

This study is limited by its small dataset size and potential scanner and acquisition heterogeneity. External validation is therefore required. Future work will evaluate skull-stripped retraining and larger multicenter cohorts.
\section{Conclusion}
Day-1 multimodal MRI fusion improved score-based discrimination for three-month functional outcome prediction after AIS, with FLAIR combined with clinical variables and ADC-derived lesion biomarkers achieving the highest cross-validated AUC. However, clinical-only modeling remained strongest for threshold-dependent metrics, emphasizing the central role of structured clinical assessment. These findings suggest that FLAIR representations and ADC lesion biomarkers provide complementary prognostic information beyond clinical variables, but require external validation and prospective evaluation before clinical translation.
\section*{Compliance with Ethical Standards}
This retrospective study used anonymized clinical data acquired as part of routine care. According to the institutional guidelines of the University of Geneva and the University Hospital of Geneva, the study is covered under existing retrospective data-use approvals, and no additional protocol-specific ethics approval was required.
The study was conducted in accordance with the principles of the Declaration of Helsinki.
\section*{Reproducibility and Acknowledgments}
This work was supported by the FORDEMS Foundation. 
The authors declare no competing interests. Data and implementation details are available upon request. 
\bibliographystyle{IEEEbib}
\bibliography{references}

@article{feigin2021global,
  author =       "V. L. Feigin \textit{et al.}",
  title =        "Global, regional, and national burden of stroke and its risk factors, 1990–2019: a systematic analysis for the Global Burden of Disease Study 2019",
  journal =      "Lancet Neurol.",
  volume =       "20",
  number =       "10",
  pages =        "795--820",
  year =         "2021",
  publisher =    "Elsevier"
}

@article{heiss2017neuroimaging,
  author =       "W. D. Heiss",
  title =        "Contribution of neuro-imaging for prediction of functional recovery after ischemic stroke",
  journal =      "Cerebrovascular Diseases",
  volume =       "44",
  number =       "5-6",
  pages =        "266--276",
  year =         "2017",
  publisher =    "Karger"
}

@article{banks2007stroke,
  author =       "J. L. Banks and C. A. Marotta",
  title =        "Outcomes validity and reliability of the modified Rankin scale: implications for stroke clinical trials: a literature review and synthesis",
  journal =      "Stroke",
  volume =       "38",
  number =       "3",
  pages =        "1091--1096",
  year =         "2007",
  publisher =    "Lippincott Williams & Wilkins"
}

@article{warach1992fast,
  author =       "S. Warach and D. Chien and W. Li and M. Ronthal and R. R. Edelman",
  title =        "Fast magnetic resonance diffusion‐weighted imaging of acute human stroke",
  journal =      "Neurology",
  volume =       "42",
  number =       "9",
  pages =        "1717--1723",
  year =         "1992",
  publisher =    "Lippincott Williams & Wilkins"
}

@article{wouters2018prediction,
  author =       "A. Wouters and C. Nysten and V. Thijs and R. Lemmens",
  title =        "Prediction of outcome in patients with acute ischemic stroke based on initial severity and improvement in the first 24 h",
  journal =      "Frontiers in Neurology",
  volume =       "9",
  pages =        "308",
  year =         "2018",
  publisher =    "Frontiers Media"
}

@article{nielsen2018prediction,
  author =       "A. Nielsen and M. B. Hansen and A. Tietze and K. Mouridsen",
  title =        "Prediction of tissue outcome and assessment of treatment effect in acute ischemic stroke using deep learning",
  journal =      "Stroke",
  volume =       "49",
  number =       "6",
  pages =        "1394--1401",
  year =         "2018",
  publisher =    "Lippincott Williams \& Wilkins"
}

@inproceedings{raeisadigh2026comparing,
  author =       "S. Raeisadigh and M. J. T. Tan and H. M{\"u}ller and A. Hedjoudje",
  title =        "Comparing Baseline and Day-1 Diffusion {MRI} Using Multimodal Deep Embeddings for Stroke Outcome Prediction",
  booktitle =    "2026 IEEE 23rd International Symposium on Biomedical Imaging (ISBI)",
  pages =        "1--5",
  year =         "2026",
  publisher =    "IEEE",
  doi =          "10.1109/ISBI61048.2026.11515670",
  url =          "https://ieeexplore.ieee.org/document/11515670"
}

@article{kim2023deep,
  author =       "D. Y. Kim \textit{et al.}",
  title =        "Deep learning-based personalised outcome prediction after acute ischaemic stroke",
  journal =      "Journal of Neurology, Neurosurgery \& Psychiatry",
  volume =       "94",
  number =       "5",
  pages =        "369--378",
  year =         "2023",
  publisher =    "BMJ Publishing Group"
}

@article{lecun2015deep,
  author =       "Y. LeCun and Y. Bengio and G. Hinton",
  title =        "Deep learning",
  journal =      "Nature",
  volume =       "521",
  number =       "7553",
  pages =        "436--444",
  year =         "2015",
  publisher =    "Nature Publishing Group"
}

@inproceedings{ma2024transformer,
  author =       "D. Ma and M. Wang and A. Xiang and Z. Qi and Q. Yang",
  title =        "Transformer-based classification outcome prediction for multimodal stroke treatment",
  booktitle =    "2024 IEEE 2nd International Conference on Sensors, Electronics and Computer Engineering (ICSECE)",
  pages =        "383--386",
  year =         "2024",
  publisher =    "IEEE"
}

@article{wold1987principal,
  author =       "S. Wold and K. Esbensen and P. Geladi",
  title =        "Principal component analysis",
  journal =      "Chemometrics and Intelligent Laboratory Systems",
  volume =       "2",
  number =       "1-3",
  pages =        "37--52",
  year =         "1987",
  publisher =    "Elsevier"
}

@article{cardoso2022monai,
  author =       "M. J. Cardoso \textit{et al.}",
  title =        "Monai: An open-source framework for deep learning in healthcare",
  journal =      "arXiv preprint arXiv:2211.02701",
  year =         "2022",
  url =          "https://arxiv.org/abs/2211.02701"
}

@article{chen2019med3d,
  author =       "S. Chen and K. Ma and Y. Zheng",
  title =        "Med3D: Transfer learning for 3D medical image analysis",
  journal =      "arXiv preprint arXiv:1904.00625",
  year =         "2019",
  url =          "https://arxiv.org/abs/1904.00625"
}

@article{cortes1995support,
  author =       "C. Cortes and V. Vapnik",
  title =        "Support-vector networks",
  journal =      "Machine Learning",
  volume =       "20",
  number =       "3",
  pages =        "273--297",
  year =         "1995",
  publisher =    "Springer"
}

@article{pedregosa2011scikit_learn,
  author =       "F. Pedregosa \textit{et al.}",
  title =        "Scikit-learn: Machine learning in Python",
  journal =      "The Journal of Machine Learning Research",
  volume =       "12",
  pages =        "2825--2830",
  year =         "2011",
  publisher =    "JMLR.org"
}
\end{document}